\documentclass[journal]{IEEEtran}
\usepackage{cite}
\usepackage{amsmath,amssymb,amsfonts}
\usepackage{mdwmath}
\usepackage{array}
\usepackage{graphicx}
\usepackage{multirow}
\usepackage{enumerate}
\usepackage{color}
\newtheorem{definition}{\textbf{Definition}}
\usepackage[ruled,linesnumbered]{algorithm2e}
\usepackage[caption=false,font=footnotesize,labelfont=sf,textfont=sf]{subfig}

\begin{document}
\title{Architecture Augmentation for Performance Predictor Based on Graph Isomorphism}

 \author{Xiangning Xie,
       Yuqiao Liu,~\IEEEmembership{Graduate Student Member,~IEEE,}
       Yanan Sun,~\IEEEmembership{Member,~IEEE}\\
       Mengjie Zhang,~\IEEEmembership{Fellow,~IEEE,}
       Kay Chen TAN,~\IEEEmembership{Fellow,~IEEE}
 \thanks{Xiangning Xie, Yuqiao Liu, Yanan Sun are with the College of Computer Science, Sichuan University, Chengdu 610065, China (e-mail: xnxie@stu.scu.edu.cn; lyqguitar@gmail.com; ysun@scu.edu.cn).}
 \thanks{Mengjie Zhang is with the School of Engineering and Computer Science, Victoria University of Wellington, Wellington, New Zealand (e-mail:mengjie.zhang@ecs.vuw.ac.nz).}
 \thanks{Kay Chen Tan is with the Department of Computing, Hong Kong Polytechnic University, Hong Kong SAR (e-mail:kctan@polyu.edu.hk).}
 }

\maketitle
\begin{abstract}
Neural Architecture Search (NAS) can automatically design architectures for deep neural networks (DNNs) and has become one of the hottest research topics in the current machine learning community. However, NAS is often computationally expensive because a large number of DNNs require to be trained for obtaining performance during the search process. Performance predictors can greatly alleviate the prohibitive cost of NAS by directly predicting the performance of DNNs. However, building satisfactory performance predictors highly depends on enough trained DNN architectures, which are difficult to obtain in most scenarios. To solve this critical issue, we propose an effective DNN architecture augmentation method named GIAug in this paper. Specifically, we first propose a mechanism based on graph isomorphism, which has the merit of efficiently generating a factorial of $\boldsymbol n$ (i.e., $\boldsymbol n!$) diverse annotated architectures upon a single architecture having $\boldsymbol n$ nodes. In addition, we also design a generic method to encode the architectures into the form suitable to most prediction models. As a result, GIAug can be flexibly utilized by various existing performance predictors-based NAS algorithms. We perform extensive experiments on CIFAR-10 and ImageNet benchmark datasets on small-, medium- and large-scale search space. The experiments show that GIAug can significantly enhance the performance of most state-of-the-art peer predictors. In addition, GIAug can save three magnitude order of computation cost at most on ImageNet yet with similar performance when compared with state-of-the-art NAS algorithms.

\end{abstract}

\IEEEpeerreviewmaketitle

\section{Introduction}\label{sec1}
\IEEEPARstart{D}{eep} Neural Networks (DNNs) gain much success in many real-world applications, spanning image classification~\cite{2016Deep, huang2017densely}, natural language processing~\cite{devlin2018bert}, and object detection~\cite{girshick2015fast, redmon2016you}, to name a few. The success of DNNs highly relies upon their novel architectures. This can be evidenced from state of the arts such as ResNet~\cite{2016Deep}, DenseNet~\cite{huang2017densely}, and Transformer~\cite{vaswani2017attention}. However, the architectures of these DNNs are often manually designed through the trial-and-error process, which is quite labor-intensive and time-consuming, and inevitably limits the diverse applications of DNNs. This greatly promotes the research of Neural Architecture Search (NAS), which can automatically design high-performance DNN architectures without or with little human expertise. On some occasions,  the architectures designed by NAS can even surpass the hand-crafted ones in some tasks~\cite{real2019regularized, zoph2018learning}.

In NAS, the architecture design is often formalized as an optimization problem, and then the automation is achieved by solving the optimization problem through well-designed search strategies. Commonly, the NAS problem is difficult to be solved because of facing multiple challenges, such as the desecrate nature, prohibitive computation cost, and also with multi-conflicting objectives~\cite{liu2020survey}. Existing search strategies for NAS mainly covers Evolutionary Computation (EC)~\cite{back1997handbook}, Reinforcement Learning (RL)~\cite{kaelbling1996reinforcement}, and gradient-based algorithms. In principle, no matter which search strategy is used, they all need to evaluate the performance of DNNs. This is because the search strategies require knowing the their current profits, which in turn effectively and efficiently guides the consequence search. However, the evaluation often involves the training of a large number of DNNs, which is a prohibitively high cost. For example, the RegularizedEvo algorithm~\cite{real2019regularized} ran on 450 GPUs for 7 days. The LargeEvo algorithm~\cite{real2017large} performed on 250 GPUs for 11 days. In practice, it is unaffordable for most researchers interested. As a result, how to accelerate the NAS evaluation process has become an important topic in the NAS community~\cite{sun2019surrogate,ning2021evaluating,ren2021comprehensive,yu2020evaluating}.

During past years, some efforts have been dedicated to solving this major issue, and existing methods can be generally divided into four different categories. They are early stopping strategy~\cite{sun2018particle,sun2018experimental,wang2018hybrid}, reduced training set~\cite{sapra2020constrained}, weight-sharing method~\cite{bender2018understanding}, and performance predictor~\cite{sun2019surrogate}. Specifically, the early stopping strategy works based on the observation that the training performance does not increase significantly within a training window, and then the training is terminated with the assumption that the performance would not increase as the training continues. As a result, the early stopping strategy consumes fewer training epochs compared to the originally assigned one, thus the whole training process is accelerated. However, this assumption would not hold because of the use of the scheduled learning rates for modern DNNs in practice, resulting in the observation in early stopping would not happen~\cite{liu2021homogeneous}. The reduced training set method uses a subset as a proxy of the training set, and then the training is performed on the subset instead of the whole training set. Because the number of training samples participating in the training is reduced, the training process is accelerated. However, how to select a proper proxy dataset sufficiently representing the whole dataset is still an open question~\cite{liu2020survey}. In practice, a common way is to construct the proxy by random sampling. However, because of the powerful fitting ability of DNNs, the obtained performance through this strategy often misleads the search strategies~\cite{liu2020survey}. The weight-sharing method is widely used by gradient-based NAS algorithms. It trains a supernet subsuming all candidate DNN architectures. Once a new architecture is generated, it directly employs the weights sharing from the supernet without any further training. This is the reason why the weight-sharing methods could accelerate the NAS. Recent works show that the shared weights often cause the rank disorder problem~\cite{yang2019evaluation}, resulting in the method having similar performance to random search~\cite{yang2019evaluation,yu2020evaluating}. In addition, the use of the weight-sharing method also heavily depends on expertise because of the construction of the supernet in advance. This clearly contrasts with the design motivation of NAS which aims at reducing the manual intervention as much as possible for architecture design.

\begin{figure}[htbp]
\centerline{\includegraphics[width=0.48\textwidth]{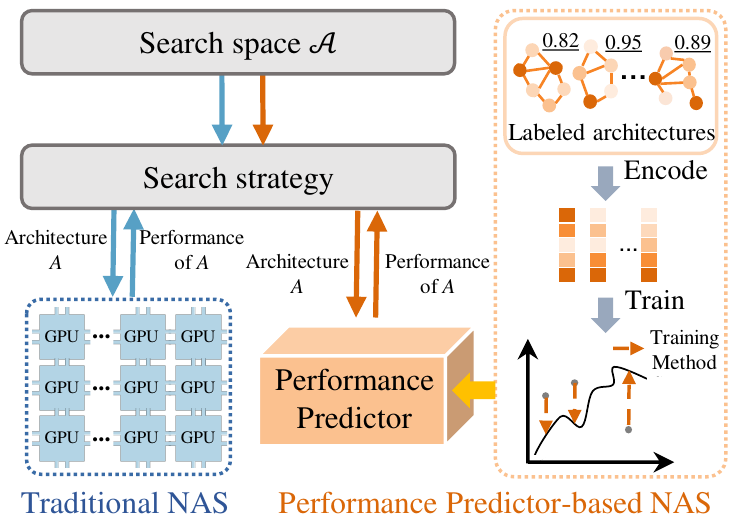}}
\caption{The illustration between traditional NAS and performance predictor-based NAS.}\label{fig_comp}
\end{figure}

The performance predictor refers to predicting the performance of DNNs using a predictor which is often a regression model instead of performing the actual evaluation. Compared to the traditional one obtaining DNN performance on a great number of Graphics Processing Units (GPUs) consuming hours to days even on medium-scale datasets, the performance predictor is of great efficiency. In practice, the prediction is often completed within seconds, and also independent of the scales of the target datasets. In addition, the performance predictor has no limitations of the early stopping strategy as well as the reducing training set method. Furthermore, there is no supernet is required for performance predictors. As a result, the performance predictor is theoretically high-fidelity and has sound merit. Fig.~\ref{fig_comp} shows the comparisons between the traditional NAS and performance predictor-based NAS. It can be seen from Fig.~\ref{fig_comp} that the performance predictor can directly estimate the performance of the searched architectures after training on a group of annotated DNN architectures. 

The first work focusing on performance predictors can be seen as the Peephole~\cite{deng2017peephole} proposed in the year of 2017. Specifically, Peephole~\cite{deng2017peephole} adopted the Long-Short Term Memory (LSTM)~\cite{hochreiter1997long} model to predict the performance, by considering that the LSTM is suitable for sequential data to which the searched DNN architectures can be processed. However, Peephole evaluated 800 DNNs as the training data of LSTM, which somehow is not suitable for real-world applications because if there are enough GPU resources to obtain the 800 annotated architectures, we would directly run the NAS algorithms on the computation resources. After that, the research efforts on performance predictors move to advanced regression models upon limited DNN architectures annotated. Sun \textit{et. al.} proposed the E2EPP performance predictor~\cite{sun2019surrogate} based on random forest~\cite{breiman2001random}, and the experiments show that E2EPP can double the running speed yet with only 200 annotated DNN architectures as training samples. 
ReNAS~\cite{xu2021renas} used a simple LeNet-5 model~\cite{lecun1998gradient} and trained it with 423 labeled architectures. Furthermore, ReNAS used the pairwise ranking loss function instead of the traditional mean squared error (MSE) because the relative ranking of DNNs is crucial for NAS.
With the use of graph convolution network (GCN)~\cite{kipf2016semi}, SSANA~\cite{tang2020semi} proposed to enhance the quality of the extracted features with an additional unsupervised model, which can learn features from unlabeled DNN architectures.
TNASP~\cite{lu2021tnasp} presented a transformer-based predictor~\cite{vaswani2017attention} and designed a self-evolution framework to fully utilize the temporal information, and achieves promising results trained with 1,000 labeled samples. 

Despite their success, there are still two limitations. \textbf{First}, these performance predictors did not solve the fundamental problem, i.e., the lacking of labeled DNN architectures. This is because the training process to obtain the performance of DNNs is time-consuming. As a result, a large number of labeled samples are not available for most researchers, and evaluating so many architectures also disobeys the designed principle of performance predictor. Although the works mentioned above proposed the advanced regression models to reduce the number of annotated architectures required, these solutions cannot solve the problem from the root. In contrast, this may lead to the problem of over-fitting due to their employment of advanced regression models but without sufficient annotated architectures~\cite{zhong2020random}. On the other hand, advanced machine learning depends on even more training data than traditional ones. Their utilization of existing performance predictors further intensifies the requirements for a large number of annotated DNN architectures. \textbf{Second}, As can be seen from Fig.~\ref{fig_comp}, the process of ``encoding'' bridges the labeled DNN architectures and the corresponding regression models. However, existing encoding methods are designed only for the specified scenario, which makes the performance predictors unable to be flexibly used by various NAS algorithms. For example, Peephole encoded the DNNs into the representation for every possible operation, and concatenate the representation of all layers as the final encoding of the architecture. However, the encoding method cannot be used when an unknown operation appears. The encoding method designed in E2EPP can only be employed for the AE-CNN NAS algorithm~\cite{sun2019completely}. TNASP designs a Laplacian matrix-based positional encoding strategy which is only designed for the input of transformer model~\cite{vaswani2017attention} and cannot be used in other models. As a result, the encoding methods of these performance predictors lack generality. However, based on the design motivation of performance predictors, they should be an independent research topic and applied by most various NAS algorithms.

In this paper, we aim to develop a Graph Isomorphism-based Architecture Augmentation method (GIAug) for designing performance predictors, which can effectively and efficiently address the above issues of existing performance predictors. Please note that this work is based on our previous preliminary investigation~\cite{liu2021homogeneous}, but with additionally significant contributions as belows: 

\begin{itemize}
	\item $\textbf{Computational cheapness.}$ The mechanism of graph isomorphism is exploited to augment DNN architectures for the first time, which can greatly benefit generating a large number of annotated architectures yet at a low cost. Specifically, all possible labeling sequences are applied to generate isomorphic graphs, resulting in various architecture representations of the same root of architectures. As a result, there is no need for intensive computation resources during this process, thus the GIAug method is computationally cheap.

	\item $\textbf{Applicable generality.}$ A generic encoding method is presented to maximally span various search spaces adopted by the NAS algorithms. This allows GIAug to be flexibly applied to different performance predictors, which can in turn enhance the efficiency of the corresponding NAS algorithms. Specifically, almost all architectures are treated as the directed attribute graph. We use the adjacency matrix to represent the vertical connection and utilize the one-hot strategy to encode the vertice attribution. This helps to be easily embedded into the performance predictor at hand.
	
	\item $\textbf{Superior effectiveness.}$ GIAug can generate sufficient training data from existing data, greatly improving the performance of the performance predictor. Furthermore, GIAug shows its superiority even in conjunction with simple regression models. Specifically, GIAug is investigated on CIFAR-10 and ImageNet datasets, with the search space of NAS-Bench-101, NAS-Bench-201, and DARTS. The results demonstrate the effectiveness of GIAug in improving the performance of the performance predictor. 
	
\end{itemize}

The remainder of this paper is organized as follows. The background and related work are reviewed in Section~\ref{sec_2}. This is followed by the details of the proposed algorithm documented in Section~\ref{sec3}. In order to validate the effectiveness and the efficiency of the proposed method, extensive experiments and the results are presented in Section~\ref{sec_4}. Finally, the conclusion and future work are shown in Section~\ref{sec_5}.

\section{Background and Related Work}\label{sec_2}
In this section, data augmentation is first briefly introduced in Subsection~\ref{relate_Aug}, and then the NAS and performance predictor are presented in Subsection~\ref{related_NAS}. Finally, the existing encoding scheme are reviewed in Subsections~\ref{relate_encoding}.

\subsection{Data Augmentation}\label{relate_Aug}
Data augmentation is a technique to increase the size of training data~\cite{shorten2019survey}, which aims at enhancing the performance of the corresponding learning algorithms. Most data augmentation methods are realized by modifying the copies from existing data or creating data from new. It has been commonly used to solve the problems caused by insufficient annotated training data in deep learning~\cite{lecun2015deep} which often relies on a large amount of training data. In general, the larger volume of training data that can be used, the better the corresponding deep learning model that can be learned. However, it may be challenging to obtain sufficient annotated data due to the prohibitive cost of collecting and labeling data. For instance, in medical image analysis tasks~\cite{esteva2017dermatologist}, such as liver lesion classification or brain scan analysis, many images are gained from computerized tomography and magnetic resonance imaging. These collections are often time-consuming, and the equipment used is expensive. In addition, the obtained images need to be labeled by medical experts, which is also labor-intensive. 

Many data augmentation methods have been proposed in the research field of computer vision and natural language processing. For example, flipping and cropping are the most popular augment techniques for computer vision~\cite{shorten2019survey}. Specifically, the flipping operation randomly flips the images horizontally to be a new image, and the cropping operation extracts a random subpatch from the image to be a new image. Furthermore, the color space transformation method~\cite{jurio2010comparison} is another exciting data augmentation method to deal with lighting biases in image recognition problems. It is achieved by altering the color distribution of images. The back-translation method~\cite{sennrich2015improving}, synonym replacement method~\cite{zhang2015character} and G-DAUG method~\cite{yang2020generative} are popular methods for data augmentation in natural language processing. Specifically, the back-translation method first translated the sequence of text into another language, and then backed into the original language to generate data with the same label as the original one. For synonym replacement, a random word was picked from a sentence and then replaced by its synonym word to generate new sentences. G-DAUG used transformer language models pre-trained to augment the most informative and diverse part of the text, and then generated synthetic texts.

Unfortunately, these existing methods cannot be applied to augment the data investigated in this paper, i.e., the DNN architectures. Specifically, the data processed by computer vision or natural language processing are often in the Euclidean space, and the corresponding augmentation methods are also designed for the Euclidean data~\cite{sun2021arctext}. For example, the flipping operation in augmenting image data works upon the fact that the pixels are represented with the measurements designed for Euclidean data. If the Euclidean distance between any two pixels changes after the augmentation, the augmented image will also be invalid. To the best of our knowledge, this is the first work focusing on augmenting DNN architectures. This is mainly because the study of performance predictors is still in its infancy.

\subsection{Neural Architecture Search (NAS) and Performance Predictor}\label{related_NAS}
NAS is composed of three consequential components: search space, search strategy, and performance estimation~\cite{elsken2019neural}. The search space refers to the collection of the candidate architectures, from which the NAS will perform the search. The search strategy corresponds to the employed optimization algorithms for the search. The performance estimation represents the way to evaluate the performance of the architectures searched. Among the three different categories of NAS algorithms, the EC-based and RL-based ones are often more time-consuming than the gradient-based ones~\cite{liu2018darts}. In principle, all types of NAS algorithms consumed a similar computation budget in terms of the three search strategies. Specifically, this bias is from the DARTS~\cite{liu2018darts} algorithm falling into the gradient-based category, which reported only several GPU days for the search upon the fact that the others often consumed hundreds of or even more GPU days. However, this difference is not caused by the gradient-based search strategy, but the performance estimation where the searched architectures in DARTS share the weights from a supernet pre-trained. This is in contrast to the other two types of NAS algorithms, where the weights are often trained from scratch. To be honest, if the search strategies are kept the same among the different types of NAS algorithms, they will consume similar computation costs.

Given the high computation cost caused by the train-from-scratch process in NAS, in addition to the weight-sharing discussed above, there are also some other representatives~\cite{sun2018particle,sun2018experimental,wang2018hybrid,sapra2020constrained}. Among those, the performance predictors have been becoming a hot topic~\cite{sun2019surrogate,wen2020neural,xu2021renas,deng2017peephole,tang2020semi}, which follow a similar protocol in constructing the predictors:

\begin{enumerate}[\textbf{Step} 1]
	\item Sample and train some architectures in the predefined search space to serve as training data;
	\item Train a performance predictor to map the encoding of architectures and the corresponding performance values;\label{np_step1}
	\item Use this trained performance predictor to estimate the performance of architectures newly generated during NAS.
\end{enumerate}

The process of training a performance predictor in \textbf{Step}~\ref{np_step1} is often modeled as a regression task. Specifically, the performance predictor is assumed to be represented by a regression model $R$. The data used to train the predictor is denoted as ${[X,y]}$, where $X=\{X_{1}, X_{2}, \ldots, X_{N}\}$ represents the DNN architectures and $y$ represents the respective performance of the DNN architectures. The training process of $R$ can be mathematically described by Equation~(\ref{eq_PP}):
\begin{equation}\label{eq_PP}
	\min _{T_{p}} \frac{1}{N} \sum_{n=1}^{N} \mathcal{L}\left(R\left(T_{p}, Encoder(X_{n})\right), y_{n}\right)
\end{equation}
where $T_{p}$ is the trainable parameters $R$, and $\mathcal{L}(\cdot)$ denotes the loss function. The trained $R$ can by directly used to estimate performance during the NAS. Please note that $Encoder(\cdot)$ denotes the encoding scheme, which is detailed in the next subsection.

\subsection{Encoding Scheme of NAS}\label{relate_encoding}
Based on the type of basic units, the existing search spaces span over the layer-based one, the block-based one, and the cell-based one~\cite{liu2020survey}. In particular, the basic unit in the layer-based search space is mainly the primitive layers of DNNs (e.g., the convolutional layer and the pooling layer), which are often adopted by the early NAS algorithms~\cite{sun2019evolving}. As for the block-based search space, the basic unit is the different combinations of primitive layers, such as the ResNet~\cite{2016Deep} block and the DenseNet~\cite{huang2017densely} block. This kind of search space can often incur better performance than the previous one in practice~\cite{sun2020automatically,wu2019fbnet}. The architectures in the cell-based search space are the repetitions of fixed structures (i.e., the cells). Each fixed structure is often the block in the block-based search space. Each cell can be considered as a micro architecture. The recent state of arts is typically based on this search space, and can often achieve a good balance between the performance and computation cost~\cite{zoph2018learning}. Upon this, there have been increasingly related works largely designed for the cell-based search space, including the encoding scheme.

The existing encoding schemes for cell-based search spaces are mainly composed of sequence-based encoding schemes and graph-based encoding schemes. The first ones typically only encode the specific serialized information of the architectures. For example, the NAO method~\cite{luo2018neural} used a string sequence, which describes different operations in the cell, to encode the whole architecture. Those encoding schemes are straightforward and easy to implement. However, the topological information cannot be fully encoded because it is hard to be represented by string sequence~\cite{white2020study}. Because the topology plays an important part in the architectures, the performance of the performance predictors built upon those encoding schemes will deteriorate. Alternatively, both the information of topology and operations can be treated by the graph-based encoding schemes. To be specific, the serialized information of each operation is encoded as an operation list, and the topological informatiosn is first modeled by a graph and then encoded as an adjacency matrix. Both are collectively used as the whole encoding. For example, ReNAS~\cite{xu2021renas} firstly obtained the vectors of the operations, the floating-point operations per second (FLOP), and the parameters for all nodes in the graph. Then, those vectors were broadcasted into the adjacency matrix to generate the type matrix, the FLOP matrix, and the parameter matrix, respectively. At last, those matrixes were concatenated as the encoding. In this paper, we utilize a simple and general graph-based encoding method that is easy to implement.

\section{The Proposed Algorithm}\label{sec3}
In this section, we first introduce graph isomorphism in Subsection~\ref{sec30}. Then we present the overall framework of the proposed algorithm in Subsection~\ref{sec31}. After that, we detail its core components in Subsections~\ref{sec32} and~\ref{sec33}, respectively.

\subsection{Graph Isomorphism}\label{sec30}
Most architectures can be treated as the directed attributed graph, where vertices correspond to operations and edges represent the connection between vertices. 
For a directed attributed graph, the definition of graph isomorphism is as the following:

\begin{definition}\label{thm1}
	If a directed attributed graph $G_{1}=\left\{V_{1}, E_{1}\right\}$ with attribute function $L_{1}$ and a directed attributed graph $G_{2}=\left\{V_{2}, E_{2}\right\}$ with attribute function $L_{2}$ are isomorphic, then there exists a bijection $f$ from the set $V_{1}$ to the set $V_{2}$ satisfying the following condition:

	(1) if $(u,v) \in E_{1}$, $(f(u),f(v)) \in E_{2}$ and $L_{1}(u) = L_{2}(f(u))$, $L_{1}(v) = L_{2}(f(v))$;

	(2) if $(f(u),f(v)) \in E_{2}$, $(u,v) \in E_{1}$ and $L_{2}(f(u)) = L_{1}(u)$, $L_{2}(f(v)) = L_{1}(v)$.
\end{definition}

Obviously, the isomorphic directed attributed graphs essentially have the same vertice connection and vertice attribution, only differ in the position of the vertice which has no influence on the performance of architecture. As a result, isomorphic architectures have the same performance value. Motivated by the observation, we give a rise to the idea that utilizing the graph isomorphism to augment the architectures.

\textbf{Architecture isomorphism detection.} Generally, we can determine if two architectures are isomorphic by identifying if their graphs are isomorphic. It is easy for us to find a finite method to judge if two graphs are isomorphic. Given two graphs $G_{1}$ and $G_{2}$ with $n$ vertices, we can separately label the vertices of $G_{1}$ and $G_{2}$ with $1,2,\dots ,n$. Two graphs are isomorphic if one can be transformed into the other one by relabeling vertices. 
In other words, the graph $G_{1}$ and graph $G_{2}$ are isomorphic if and only if a permutation matrix $P$ exists such that $A_{2}=PA_{1} P^{-1}$ and $L_{2}=L_{1}P^{-1}$. 
Specifically, $A_{1}$ and $A_{2}$ define the adjacency matrix of $G_{1}$ and $G_{2}$. The adjacency matrix can represent the connection of the graph by its adjacency matrix $A$, defined as:
\begin{eqnarray}
	A_{i j} & = & \left\{\begin{array}{ll}
	1 & \text { if edge } (i \rightarrow j) \text { exists }  \\
	0 & \text { if edge } (i \rightarrow j) \text { does not exist}
	\end{array}\right.
\end{eqnarray}
Each adjacency matrix is built on the basis of the given labeling of each vertice. For example, a vertice with labeling $l$ indexes at $l$-th row and $l$-th column.
Furthermore, the attribute vectors $L_{1}$ and $L_{2}$ represent the attribute of all vertices in $G_{1}$ and $G_{2}$, respectively. Concretely, the $i$-th element in $L_{1}$ is the attribute type of the vertice labelled $i$. The adjacency matrix and attribute vector correspond to the topological and attribute information, respectively. 
Finally, $P$ represents the permutation matrix which is a square binary matrix that has exactly one entry of 1 in each row and each column and 0 elsewhere. The permutation matrix represents the relabeling of the elements. When it is used to multiply another matrix $M$, it results in permuting the rows (when pre-multiplying, to form PM) or columns (when post-multiplying, to form MP) of the matrix $M$. As a result, we can judge if one graph can be transformed by relabeling another graph by using the permutation matrix to multiply.

\begin{figure*}[htbp]
\centerline{\includegraphics[width=1.0\textwidth]{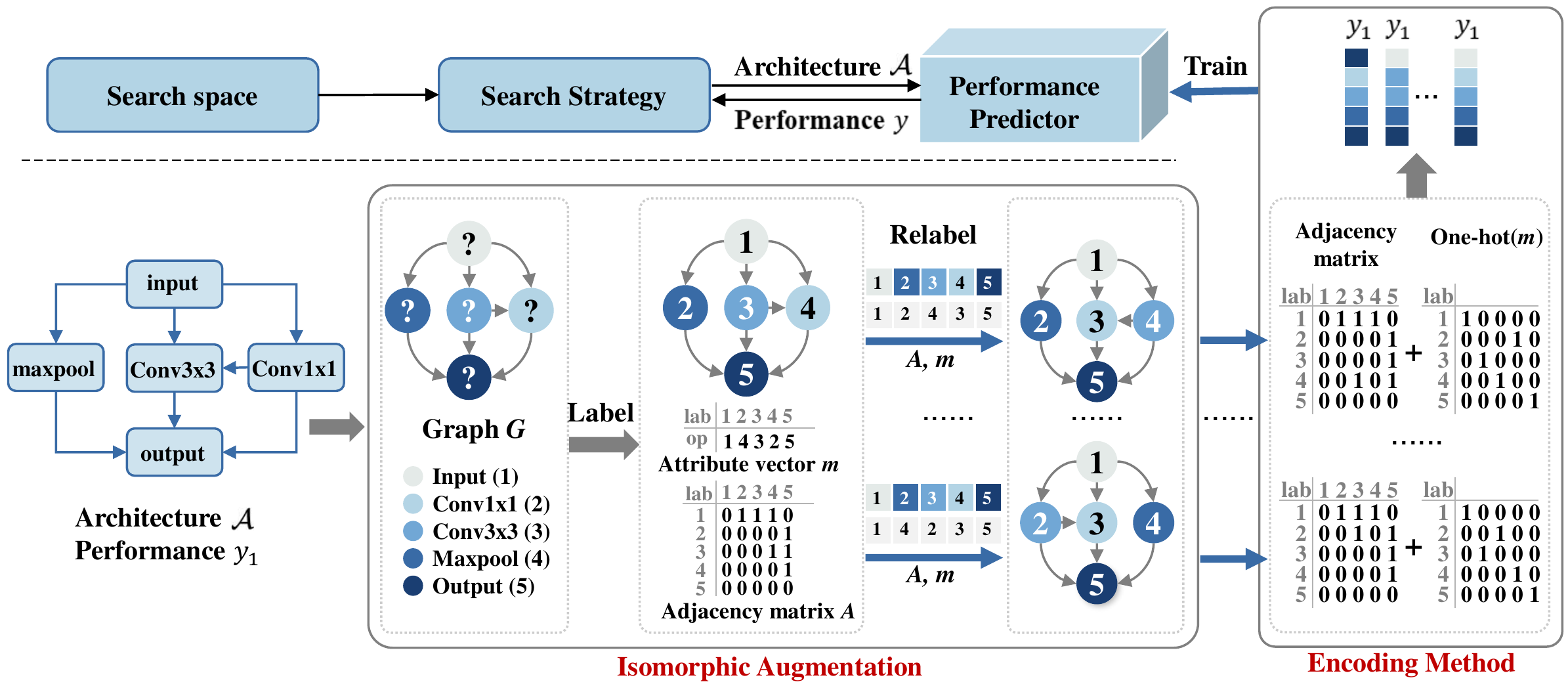}}
\caption{The flowchart of the proposed GIAug. The number of the vertices means the label, while the color of vertices corresponds to the operation. Graph $G$ and all isomorphic graphs represent the architecture $\mathcal{A}$.}\label{fig_overall}
\end{figure*}

\subsection{Overall Framework}\label{sec31}
\begin{algorithm}
\caption{Overall framework}\label{alg_overall}
\KwIn{The architecture dataset $D$.}
\KwOut{The encoded architecture dataset $D'$ after data augmentation.}
$D' \leftarrow \emptyset$ \;
\ForEach {Architecture $\mathcal{A}$ in $D$}
{\label{alg1_main_start}
$G \leftarrow$Build a graph based on $\mathcal{A}$\;\label{alg1_main_line1}
$V=(v_{1}, v_{2},\dots ,v_{n}) \leftarrow$Construct a vertice sequence based on a random topological order\;\label{alg1_main_line2}
$r=(1,2,\dots ,n) \leftarrow$Build a label sequence\;\label{alg1_main_line3}
$A \leftarrow$Construct an adjacency matrix based on $G$, $V$ and $r$\;\label{alg1_main_line4}
$m \leftarrow$ Obtain the attribute vector of $V$\;\label{alg1_main_line5}
$S \leftarrow$Permutate the labeling in $r$ to get all the possible labeling sequences\;\label{alg1_main_line6}
\ForEach {$labeling$ $sequence$ $l$ in S}{
$G' \leftarrow$Get the graph based on \textbf{the proposed isomorphic augmentation method} by $A$, $m$ and $l$\;\label{alg_main_aug}
$\mathcal{E} \leftarrow$Encoding $G'$ by \textbf{the encoding method}\;\label{alg_main_encoding}
\If{$\mathcal{E} \notin D'$}{\label{alg_if}
$D' \leftarrow D'\cup \mathcal{E}$\;} \label{alg_ifnot}
}
}\label{alg1_main_end}
$\textbf{Return}$ $D'.$\label{alg1_end}
\end{algorithm}
Algorithm~\ref{alg_overall} shows the framework of the proposed algorithm. Given the dataset $D$ containing a limited number of annotated DNN architectures, the proposed GIAug algorithm starts to take effect for the architectures in $D$ one by one (Lines~\ref{alg1_main_start}-\ref{alg1_main_end}). Finally, a new dataset $D'$ containing the encoding information of all the augmented architectures is returned for the use of the performance predictor (Line~\ref{alg1_end}). During the process of each architecture $\mathcal{A}$ in $D$, the graph of $\mathcal{A}$, say $G$, is first built by extracting the topological and attribute information (Line~\ref{alg1_main_line1}), and then a vertice sequence $V$ based on a random topological order is obtained (Line~\ref{alg1_main_line2}). Specifically, the topological order is a linear ordering of its vertice, where vertice $u$ comes before vertice $v$ for every edge $(u,v)$ from $u$ to $v$. Next, the labeling sequence $r$ is created to label each vertice in $V$ with $r_{1},r_{2},\dots, r_{N}$ (Line~\ref{alg1_main_line3}). After that, the adjacency matrix $A$ and attribute vector $m$ are constructed based on $G$, $V$ and $r$ (Lines~\ref{alg1_main_line4}-\ref{alg1_main_line5}). Please note that each operation is represented as an integer. Next, all the possible labeling sequences of $V$ are obtained by permuting the elements in $r$ and then stored into $S$ (Line~\ref{alg1_main_line6}).
At last, each graph $G'$ is generated based on the proposed isomorphic augmentation method (Line~\ref{alg_main_aug}), and its information is encoded to $\mathcal{E}$ by the proposed encoding method (Line~\ref{alg_main_encoding}). If $\mathcal{E}$ not exists in $D'$, it will be stored in $D'$ (Lines~\ref{alg_if}-\ref{alg_ifnot}). 

Fig.~\ref{fig_overall} shows an example of the proposed GIAug algorithm. Specifically, the example architecture (i.e., Architecture~$\mathcal{A}$) is composed of five nodes and five operations (i.e., input, output, max pooling, 1$\times$1 convolution, and 3$\times$3 convolution). The architecture $\mathcal{A}$ is first represented as a directed attributed graph $G$. Then, all vertices in $G$ are labeled, and the adjacency matrix $A$ and the attribute vector $m$ are obtained based on the labeling. Then, the Isomorphic Augmentation method is used to gain all isomorphic graphs of $G$ by relabeling $G$. At last, each isomorphic graph is represented to a unique encoding, and all encodings are utilized to train the performance predictor to improve its prediction ability. 


\subsection{The Proposed Isomorphic Augmentation Method}\label{sec32}
As discussed above, the proposed augmentation method is motivated by the isomorphism in the graphs, which is defined as Definition~\ref{thm1}.
In the isomorphic graphs, the relationships between vertices and the attribution of vertices in each graph keep unchanged, but the labels of the vertices are different. Based on this design motivation, the isomorphism of the graph can be utilized to augment DNN architectures for performance predictors because the topological and attribute information is crucial to such kinds of tasks, while the labeling of the vertices are meaningless. For each given annotated architecture $\mathcal{A}$ to be augmented, all its isomorphic graphs will have the same performance values as that of itself. As a result, the performance predictor can have a large number of annotated architectures. Obviously, this process is computationally cheap.

Algorithm~\ref{algorithm1} shows the details of the proposed augmentation method based on the isomorphism. Generally, the proposed isomorphic augmentation method is composed of two parts: the isomorphic information part (Lines~\ref{alg2_p1_start}-\ref{alg2_p1_end}), the graph generation part (Lines~\ref{alg2_p2_start}-\ref{alg2_p2_end}). The first part, as shown by its name, is mainly for the generation of isomorphic information. Specifically, based on the labeling sequence $l$, the permutation matrix $P$ is created to relabel the vertices (Lines~\ref{alg2_p1_start}-\ref{alg2_p1_p_gene}). Then, the adjacency matrix and the attribute vector after relabeling are calculated by multiplying the permutation matrix $P$. Thus, the topological information (i.e., the adjacency matrix) and the attribute information (i.e., the attribute vector) of the isomorphic graph to be generated are obtained. The third part is designed for the generation of the graph $G'$ using the above available information. In particular, the vertices $V'$ are generated by the adjacency matrix $A'$ (Lines~\ref{create_vertice_begin}-\ref{create_vertice_end}) and the edges $E'$ are built by the attribute vector $m'$ (Lines~\ref{create_edge_begin}-\ref{create_edge_end}). At last, the isomorphic graph $G'$ is created by $V'$ and $E'$. Note that the labeling of the input vertice and output vertice keep unchanged, because the operations of input and output are placeholders and meaningless for the performance of DNNs in performance predictors.

\begin{algorithm}
	\caption{The Proposed Augmentation Method}\label{algorithm1}
	\KwIn{The adjacency matrix $A$, the attribute vector $m$, and the labeling sequence $l$.}
	\KwOut{The isomorphic graph $G'$.}
	$N \leftarrow$ Count the number of vertices\;\label{alg2_p1_start}
	$P \leftarrow$ Create a empty matrix with size of (N,N)\;
	$i = 0$\;
	\For{$j$ in $l$}{$P_{ij}=1$\;$i \leftarrow i+1$\;}\label{alg2_p1_p_gene}
	$A' = PAP^{-1}$\;
	$m' = mP^{-1}$\;\label{alg2_p1_end}
	$V' \leftarrow \emptyset$\;\label{alg2_p2_start}
	$E' \leftarrow \emptyset$\;
	\For {i $\leftarrow$ 1 to $N$}{\label{create_vertice_begin}
		$v_i \leftarrow$ Create vertice with attribute $m'_{i}$ \;\label{alg2_p2_l1}
		$V' \leftarrow V'\cup v_i$\;
	}\label{create_vertice_end}
	\For {i $\leftarrow$ 1 to $N$}{ \label{create_edge_begin}
		\For {j $\leftarrow$ 1 to $N$}{
			\If{$A'_{ij}==1$}{
			$e' \leftarrow (V'_{i}, V'_{j})$\;
			$E' \leftarrow E'\cup e'$\;
			}
		}
		}\label{create_edge_end}
	$G' \leftarrow \{V',E'\}$\;\label{alg2_p2_end}
	$\textbf{Return}$ $G'.$
\end{algorithm}

The proposed augmentation method has the following merits. Firstly, it is computationally friendly and there is no need for intensive computation sources to train the augmented DNN architectures for annotation. This is contributed by the isomorphism that could assign the same label of the base architecture to all the augmented architectures. As a result, the proposed augmentation method does not need extra computational resources for the annotation as usual. 
Secondly, it can efficiently improve the performance of the performance predictors by generating sufficient data, which in turn enhances the superiority of the DNN architectures designed by the corresponding NAS algorithms. Specifically, the proposed augmentation method is expected to enhance the generalization ability of the performance predictor, thus improving the prediction ability on unseen architectures. The ablation study will prove the ability of the proposed augmentation method and can be checked from Subsection~\ref{exp_ablation_study}.

\subsection{The Encoding Method}\label{sec33}

The augmented architectures in the proposed algorithm are represented by graphs,  which cannot efficiently serve as the input of performance predictors, although the graph-based models, such as GNN, can directly process this kind of data structure. The main reason is that graph-based models typically use average methods to accumulate information from each vertice in the graph, and the resulting information cannot be very descriptive~\cite{wu2020comprehensive}. As a result, the performance predictor will deteriorate. As suggested by one of our recent work~\cite{sun2021arctext}, the nature language-like description of DNNs would be more suitable for performance predictors in the context of NAS. Therefore, we propose the encoding method for this purpose. However, different types of architecture are independent of encoding, which requires the designed encoding method to be generic. As discussed in Section~\ref{sec_2}, the existing DNN architectures in NAS (i.e., the search space) can be classified into Operation On Node (OON) and Operation On Edge (OOE). Specifically, for the OON-based DNN architectures, the vertices represent the operation (e.g., 3$\times$3 convolution, 3$\times$3 max-pooling) and the edges represent the connections between vertices. In the OOE-based DNN architectures, the edges represent the operations, and the vertices represent the connections between the operations. In this paper, the OOE-based architectures are transformed to the OON-based architectures following the convention~\cite{liu2021homogeneous}. Furthermore, we propose a simple yet efficient encoding method for the OON-based architectures. 

The proposed encoding method is simple and efficient. First, the edges of the graph $G$, representing the topological information of architecture $\mathcal{A}$, are described by an adjacency matrix $A \in {\{0,1\}}^{|V|\times |V|}$ initialized with all zeros. Then, the value of the element at the interaction of $i$-th row and $j$-th column of $M_A$ is specified as one if vertice with labeling $i$ and vertice with labeling $j$ are connected. Next, the attribute vector $m$ of $G$, in which the $i$-th element means the operation of the vertice labeled $i$, is transformed into an attribute matrix with the one-hot method. Finally, the matrix $M_A$ and the one-hotted attribute matrix are concatenated as the encoding of architecture $\mathcal{A}$. Particularly, the last two steps can be summarized by Equation~(\ref{equ_conct_oon}):
\begin{equation}\label{equ_conct_oon}
	E(A,m) = concat\{flatten(A),flatten(oh(m))\}
\end{equation}
where $m$ denotes the attribute vector of $G$, $oh(\cdot)$ denotes the one-hot method, and $flatten(\cdot)$ and $concat(\cdot)$ denote the flatting of the matrix and the concatenation of vectors, respectively. Specifically, the adjacency matrix $A$ is flattened into a one-dimension vector. The attribute matrix is also flattened into a one-dimension vector after the use of the one-hot method, and the concatenation is only to connect both vector strings.

\begin{figure}[htbp]
	\centerline{\includegraphics[width=0.5\textwidth]{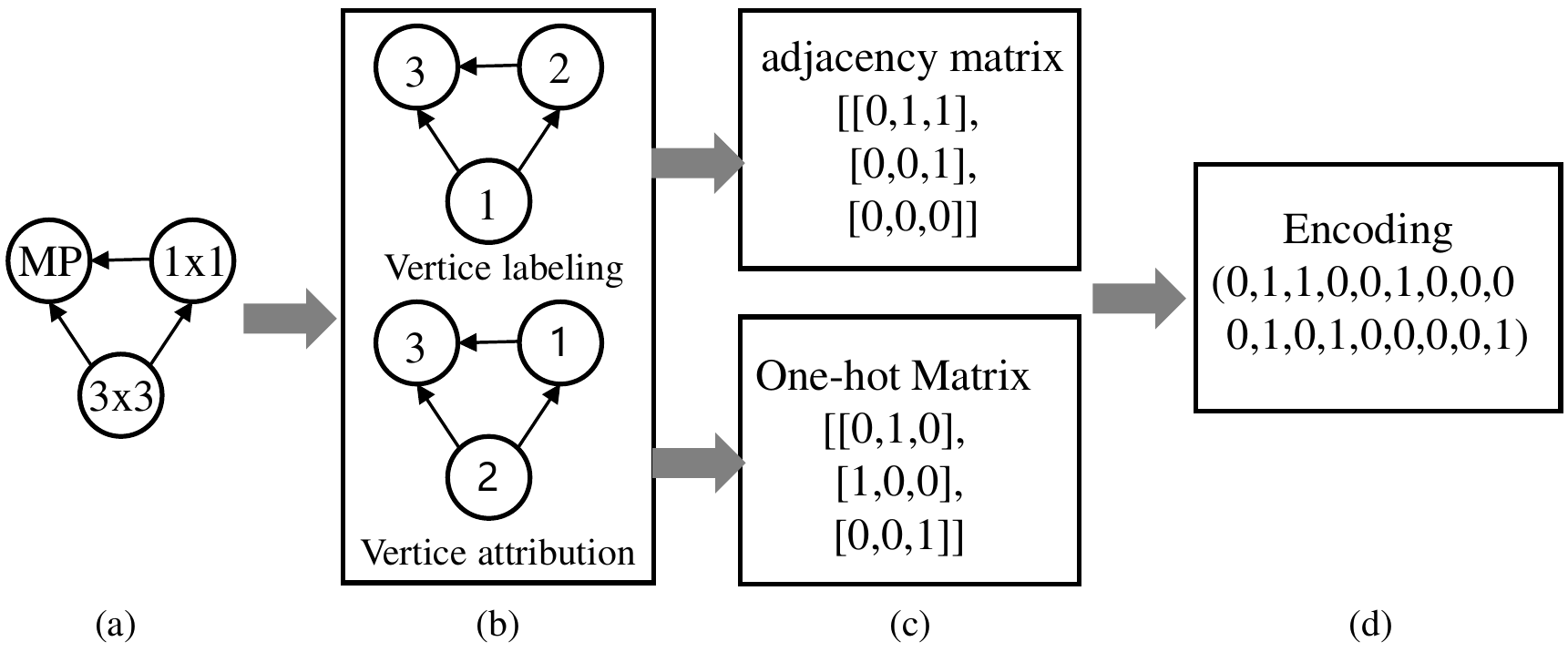}}
	\caption{An example of the encoding for OON-based architecture. \textbf{(a)}: the architecture to be represented. \textbf{(b)}: The top section shows the label of these vertices, and the bottom section shows the operation of these vertices. \textbf{(c)}: The top section is the adjacency matrix, and the bottom section is the one-hot vector of the operation list. \textbf{(d)}: the resulted encoding.}\label{fig_example_encoding_oon}
\end{figure}

An example of this encoding process is shown in Fig.~\ref{fig_example_encoding_oon} for a quick glance. Specifically, Fig.~\ref{fig_example_encoding_oon}(a) represents an OON-based architecture to be encoded, which is composed of three operations (a max-pooling vertice, two convolutional vertices with kernel sizes of $1\times 1$ and $3\times3$). In Fig.~\ref{fig_example_encoding_oon}(b), the upper denotes its labeling sequence, and the lower denotes its operation type. Based on this, its adjacency matrix and the attribute matrix transformed by the one-hot method are obtained, which are shown in the upper part and the lower part of Fig.~\ref{fig_example_encoding_oon}(c), respectively. Finally, both are connected as the whole encoding of this architecture, as shown in Fig.~\ref{fig_example_encoding_oon}(d).

\section{Experiments and Analysis}\label{sec_4}
To verify the effectiveness of the proposed GIAug algorithm, a series of experiments are conducted and then be analyzed in this section. Specifically, the search space and the corresponding datasets on which GIAug will be investigated are introduced in Subsection~\ref{sec_search_space}, and then the experiment results are demonstrated in Subsections~\ref{sec4_com_np} to \ref{exp_ablation_study}. In particular, the experiments of GIAug are conducted from three aspects: the comparison against state-of-the-art performance predictors, the comparisons against state-of-the-art NAS algorithms, and the ablation study against itself.

\begin{figure}[!htp]
	\centering
	\subfloat[Cells in NAS-Bench-101]{\includegraphics[width=0.4\textwidth]{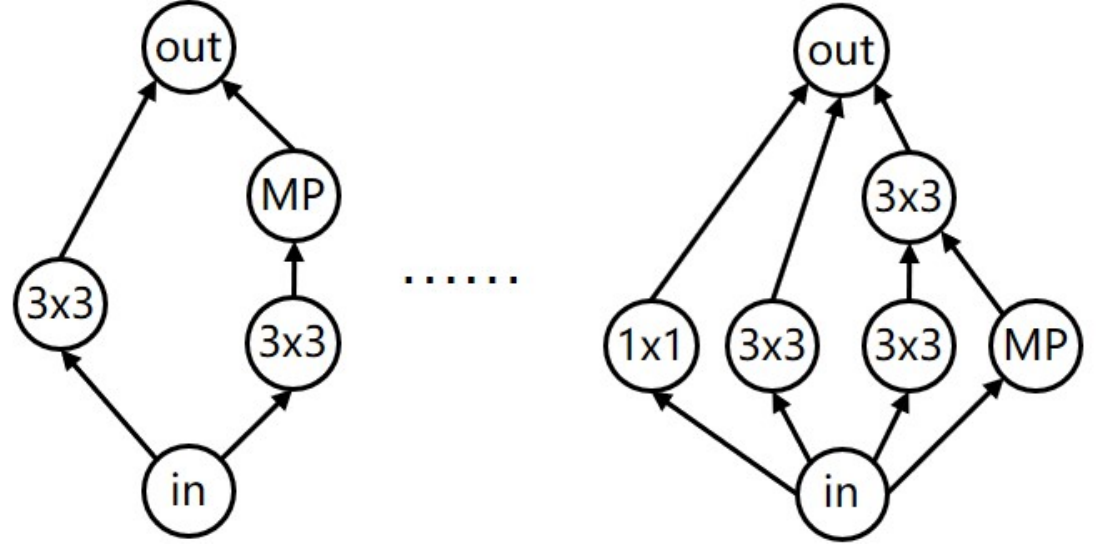}\label{fig_nas101_example}}
	\hfil
	\subfloat[Cells in NAS-Bench-201]{\includegraphics[width=0.5\textwidth]{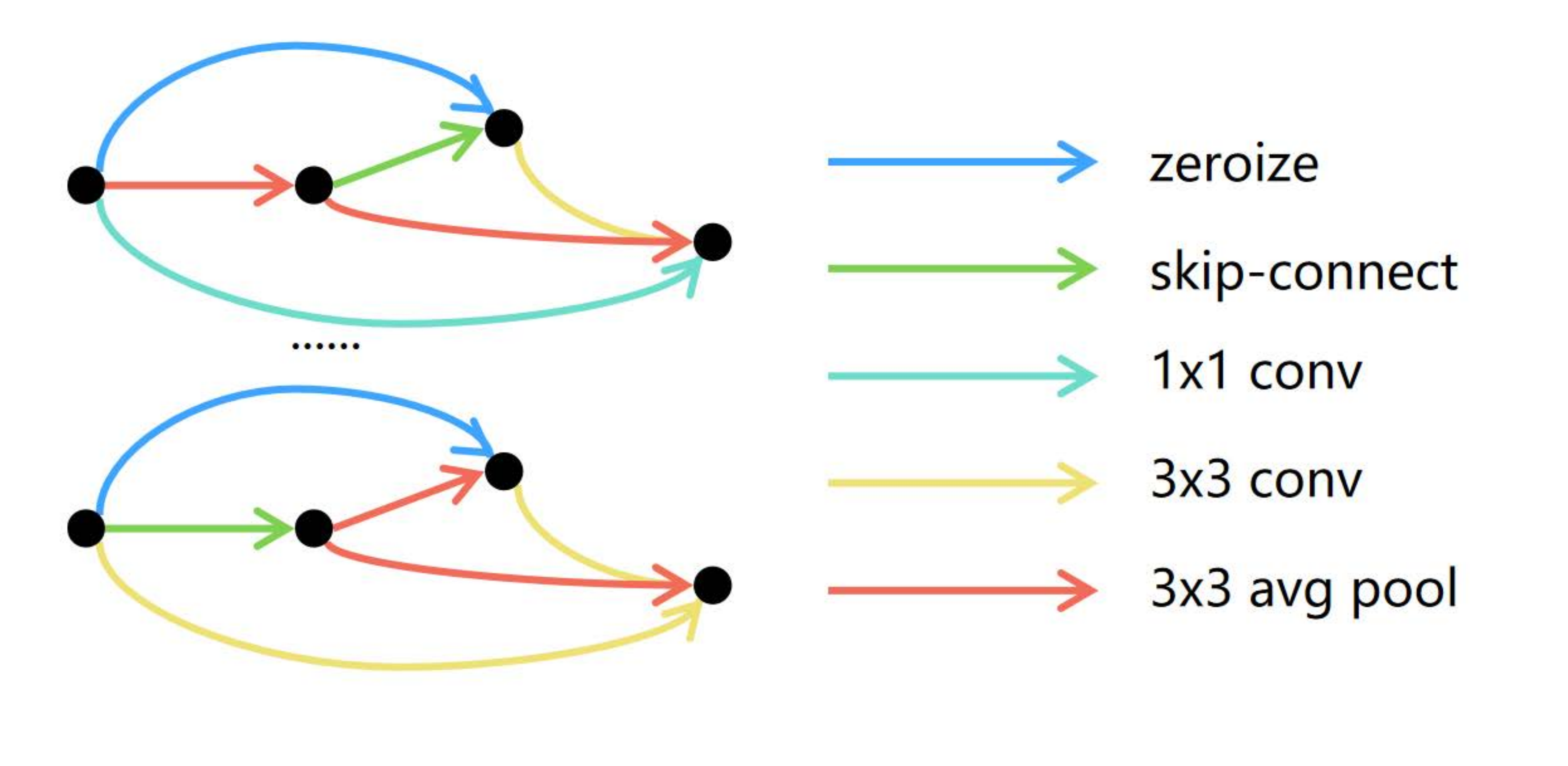}\label{fig_nas201_example}}
	\hfil
	\subfloat[Cells in DARTS]{\includegraphics[width=0.4\textwidth]{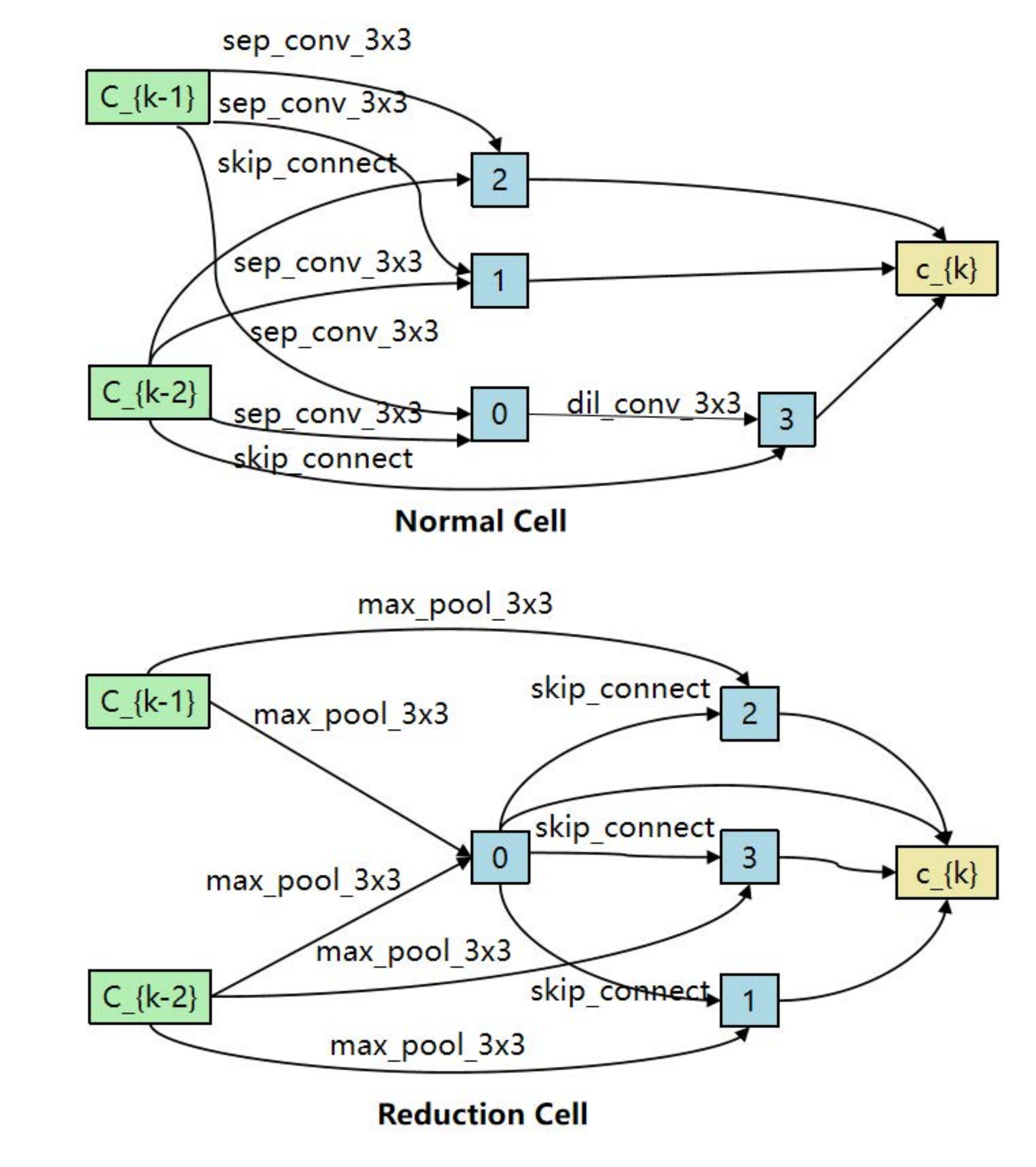}\label{fig_darts_example}}
	\caption{The cells in the search spaces NAS-Bench-101~\cite{ying2019bench}, NAS-Bench-201~\cite{dong2020bench}, and DARTS~\cite{liu2018darts}.}\label{fig_benchmark_examples}
\end{figure}

\subsection{Search Space}\label{sec_search_space}
As highlighted, existing search spaces are composed of two different categories: the OON and the OOE. In order to extensively demonstrate the superiority and the generalization of the proposed GIAug method, we perform the experiments on both types of search spaces. In this experiment, we adopt the search space of two widely used NAS benchmarks (i.e., the NAS-Bench-101~\cite{ying2019bench} and the NAS-Bench-201~\cite{dong2020bench} ) and the search space of a famous NAS algorithm (i.e., the DARTS algorithm~\cite{liu2018darts}). Specifically, the search space of NAS-Bench-101 follows the OON category, and the search spaces of the NAS-Bench-201 and DARTS follow the OOE category.

\textbf{NAS-Bench-101} is a popular benchmark dataset for NAS algorithms. It provides a large number of CNN architectures that have been exhaustively enumerated and then trained in advance. As a result, the related researchers can computationally cheaply perform experiment comparisons with their own NAS algorithms. Specifically, NAS-Bench-101 released $423,624$ different CNN architectures with performance trained on CIFAR-10 image classification benchmark dataset.

For the search space of NAS-Bench-101, it defines one 3$\times$3 convolution as the initial layer. Then it stacks each cell for three times, followed by a downsampling layer. At last, it repeats this pattern for three times, followed by global average pooling and a dense softmax layer. In addition, each cell in NAS-Bench-101 is limited to seven operations, including both the input and the output, and the possible operations are 1$\times$1 convolution, 3$\times$3 convolution, and the 3$\times$3 max-pooling. Fig.~\ref{fig_nas101_example} shows two examples of the architectures sampled from the search space of NAS-Bench-101, which has five operations and seven operations, respectively. As can be seen from this example, the operations are in the vertices, which follows the category of OON search space.

\textbf{NAS-Bench-201} is also a benchmark dataset similar to NAS-Bench-101 with the difference regarding the OOE search space, containing $15,625$ CNN architectures. Specifically, the skeleton of NAS-Bench-201 is initiated with one 3$\times$3 convolution and a batch normalization layer. The main body of the skeleton includes three stacks of cells, connected by a residual block and ends up with a global average pooling layer. Each cell has four vertices and does not limit the number of edges. Specifically, each edge of NAS-Bench-201 is associated with an operation selected from a predefined operation set including: zeroize, skip connection, 1$\times$1 convolution, 3$\times$3 convolution, an 3$\times$3 average pooling. Unlike NAS-Bench-101, the operations are defined on the edges as shown in the Fig.~\ref{fig_nas201_example}.

\textbf{DARTS} is a famous NAS algorithm based on the gradient descent category. Owing to the effectiveness in terms of the gradient-based algorithms, there have been a large number of DARTS-based variants~\cite{xu2019pc} that use the same search space as that of DARTS. In addition, in contrast to NAS-Bench-101 and NAS-Bench-201 that are mainly investigated on median-scale image classification tasks (e.g., CIFAR-10), the variants based on DARTS are often investigated on the ImageNet~\cite{deng2009imagenet} that is a challenging large-scale image dataset. Specifically, DARTS consists of two types of cells: the normal cell and the reduction cell. All normal cells are constrained to have the same architectures, as the same situations for reduction cells, but the architecture of the normal cells is independent of that of the reduction cells. In addition, every application in the reduction cells is followed by a convolutional operation with the stride of $2$ to reduce the image size, whereas normal cells keep the image size unchanged.

As shown in Fig.~\ref{fig_darts_example}, the normal cell and the reduction cell are with the same two inputs and one output. Each consists of seven vertices, and each edge in the cell is associated with an operation selected from a predefined operation set containing 3$\times$3 separable convolutions, 5$\times$5 separable convolutions, 3$\times$3 dilated separable convolutions, 3$\times$3 max pooling, 3$\times$3 average pooling, identity, and zeroize.

\begin{table*}[htp]
	\caption{The results \textbf{(measured by Kendall's Tau)} of GIAug on \textbf{NAS-Bench-101} against chosen performance predictors using training data with proportions of $0.01\%$, $0.03\%$, $0.05\%$, $0.07\%$, $0.1\%$, $0.3\%$, $0.5\%$, $0.7\%$, and $1\%$.}
	\setlength\tabcolsep{3pt}
	\begin{center}
		\begin{tabular}{c|ccccccccc}
			\hline
			\multirow{1}{*}{} 
			& 0.01\%(\%) & 0.03\%(\%) & 0.05\%(\%) & 0.07\%(\%) & 0.1\%(\%) & 0.3\%(\%) & 0.5\%(\%) & 0.7\%(\%) & 1\%(\%) \\ \hline
			AlphaX~\cite{wang2019alphax}                        & 12.17 & 32.29 & 32.96 & 36.42 & 41.09 & 41.13& 45.00 & 43.11& 44.86 \\
			Peephole~\cite{deng2017peephole}                   & 10.65 & 7.35 & 19.74 & 15.33 & 14.14 & 11.41& 22.39 & 37.50& 28.47 \\
			ReNAS~\cite{xu2021renas}                      &   -    &    -   &    -   &    -   & 65.74 &   -   &   -    &   -   & \textbf{81.61} \\
			RFGIAug(ours)                     & \textbf{46.99$\pm$0.31} & \textbf{63.42$\pm$0.33} & \textbf{65.00$\pm$0.17} & \textbf{66.58$\pm$0.14} & \textbf{67.27$\pm$0.30} & \textbf{72.36$\pm$0.10}& \textbf{74.21$\pm$0.07} & \textbf{74.86$\pm$0.05}& 76.41$\pm$0.03 \\ \hline\hline
			E2EPP~\cite{sun2019surrogate}                      & 17.59 & 37.60& 48.98& 52.84 & 52.91 & 64.98& 67.06 & 68.21& 68.82 \\
			E2EPP+GIAug(ours)                 & \textbf{48.35$\pm$0.48} & \textbf{66.29$\pm$0.29} & \textbf{65.98$\pm$0.25} & \textbf{66.76$\pm$0.27} & \textbf{67.58$\pm$0.29} & \textbf{71.60$\pm$0.32}& \textbf{72.21$\pm$0.42} & \textbf{72.71+0.15}& \textbf{73.05$\pm$0.40} \\\hline\hline
			NPNAS~\cite{wen2020neural}                      & 53.06 & 63.77 & 66.29 & 63.36 & 65.47 & 67.40& 68.36 & 70.13& 71.46 \\
			NPNAS+GIAug(ours)               & \textbf{55.67$\pm$4.62} & \textbf{64.59$\pm$3.99} & \textbf{66.48$\pm$1.27} & \textbf{67.85$\pm$0.70} & \textbf{71.04$\pm$0.30} & \textbf{73.71$\pm$0.27} & \textbf{75.34$\pm$0.27}& \textbf{76.24$\pm$0.34} & \textbf{76.74$\pm$0.15} \\ \hline
		
		\end{tabular}
	\end{center}\label{result_pp_101_ktau}
\end{table*}

\subsection{Comparison Against performance predictors}\label{sec4_com_np}

This comparison is conducted on NAS-Bench-101 and NAS-Bench-201, and two popular metrics, i.e., Kendall's Tau~\cite{sen1968estimates} and $N@K$~\cite{ning2020generic}, are used for the quantitative measurement.

Specifically, Kendall's Tau describes the correlation between the orders regarding the predicted value and the ground-truth values, which can be described by Equation~(\ref{eq_Ktau}):
\begin{equation}\label{eq_Ktau}
	\text { Kendall's Tau }=2 \times \frac{\text { number of concordant pairs }}{n(n-1) / 2}-1
\end{equation}
where $n$ denotes the number of samples, and the concordant pair means that the rankings of predicted values and the actual values of a given pair are the same. The value of Kendall's Tau ranges between $[-1, 1]$. The closer the value to $1$, the better performance of the predictors. Kendall's Tau has been widely used by existing performance predictors for the comparison~\cite{sun2019surrogate,wen2020neural,xu2021renas,deng2017peephole,tang2020semi}. However, all concordant pairs in Kendall's Tau are treated equally, which also include the poor architectures that are often out of the concerns of NAS algorithms in practice. In complement to this, the $N@K$ metric is also collectively used, which measures the best rank of the top $K$ out of the $N$ architectures predicted by the performance predictors. For example, if there are $N$ architectures in the test dataset, when the performance predictor has been built, the performance of these $N$ architectures are predicted, and then the top $K$ architectures with the highest performance are selected. After that, the ranks of these $K$ architectures with their true performance in the test dataset are collected, and then the best one is selected for the comparison. The smaller the $N@K$ value, the better the performance predictor. The $N@K$ metric is very popular among related works published recently~\cite{ning2020generic,chen2021not}.

\subsubsection{\textbf{Comparisons of Predictors on NAS-Bench-101}}\label{sec_comp_pp_101}
On the NAS-Bench-101 dataset, the state-of-the-art performance predictors chosen for the comparison are AlphaX~\cite{wang2019alphax}, Peephole~\cite{deng2017peephole}, E2EPP~\cite{sun2019surrogate}, ReNAS~\cite{xu2021renas} and NPNAS~\cite{wen2020neural}. Please note that AlphaX used different regression models in its seminal paper for the experiment demonstration, in this paper, we use its implementation based on the MLP for the comparison by following the conventions in~\cite{ning2020generic,chen2021not}. In addition, we implement the proposed GIAug method by using the Random Forest (RF)~\cite{breiman2001random} as the regression model, and named as RFGIAug for the discussion. Specifically, the RF is an ensemble model that often can easily achieve the best performance without too much manual tuning, and is very popular among various challenge competitions (the effectiveness of using RF has also been verified in the ablation study shown in Subsection~\ref{exp_ablation_study}). Furthermore, in order to show the flexibility of the proposed GIAug method, we also embed GIAug into E2EPP and NPNAS for the comparisons, which are named E2EPP+GIAug and NPNAS+GIAug, respectively, for the convenience of the discussion.

By following the conventions of the community~\cite{ning2020generic,chen2021not}, we also use $0.05\%$ and $0.1\%$ of the whole training dataset for the training of the compared performance predictors, and then test them on the remaining dataset for reporting the experiment results in terms of the chosen metrics. In addition, in order to comprehensively investigate the effectiveness of the proposed GIAug method, we additionally add another seven groups with different proportions of training data for the comparison in terms of the Kendall's Tau metric (i.e., $0.01\%$, $0.03\%$, $0.07\%$, $0.3\%$, $0.5\%$, $0.7\%$, and $1\%$). This addition is with the fact that a large amount of data is not available in practice, and this comprehensive investigation can fully reveal the effectiveness of GIAug. 
To validate the stability of GIAug, we randomly sample a specified proportion of data from NAS-Bench-101 as training data in each experiment, and repeat the experiment for $10$ times.
In addition, the number of trees in the RF is specified as $230$, and the mean squared error criterion is used as the loss function.

\begin{table}[htp]
	\caption{The results (measured by $N@K$ with $K=5$ and $10$) of GIAug on \textbf{NAS-Bench-101} against chosen performance predictors using training data with proportions of $0.05\%$ and $0.1\%$.}
	\begin{center}
		{
		\begin{tabular}{c|cccc}
			\hline
			\multirow{2}{*}{} & \multicolumn{2}{c|}{0.05\%} & \multicolumn{2}{c}{0.1\%} \\ \cline{2-5} 
			& N@5     & \multicolumn{1}{c|}{N@10}    & N@5     & N@10    \\ \hline
			AlphaX~\cite{wang2019alphax}                       & -       & \multicolumn{1}{c|}{-}       & 57      & 58      \\
			Peephole~\cite{deng2017peephole}                & 7794     & \multicolumn{1}{c|}{7794}    & 17288     & 17288    \\
			RFGIAug (Ours)                        & \textbf{24}     & \multicolumn{1}{c|}{\textbf{24}}     & \textbf{14}     & \textbf{14}        \\ \hline\hline
			
			E2EPP~\cite{sun2019surrogate}                   & 2430      & \multicolumn{1}{c|}{600}      & 384     & 384    \\
			E2EPP+GIAug (Ours)                  & \textbf{311}      & \multicolumn{1}{c|}{\textbf{62}}    & \textbf{2}      & \textbf{2}    \\\hline\hline
			NPNAS~\cite{wen2020neural}                      & 22570    & \multicolumn{1}{c|}{5279}     & 19487    & 19487    \\
			NPNAS+GIAug (Ours)                  & \textbf{574}      & \multicolumn{1}{c|}{\textbf{285}}     & \textbf{107}      & \textbf{107}    \\\hline 
		\end{tabular}}
	\end{center}\label{result_pp_101_NK}
\end{table}


\begin{table*}[htp]
   
	\caption{The results \textbf{(measured by Kendall's Tau)} of GIAug on \textbf{NAS-Bench-201} against chosen performance predictors using training data with proportions of $0.5\%$, $1\%$, $5\%$, $10\%$, $15\%$, $20\%$.}
	\begin{center}
		\begin{tabular}{c|cccccc}
			\hline
			\multirow{1}{*}{} 
			& 0.5\%(\%)       & 1\%(\%)       & 5\%(\%)       & 10\%(\%)      & 15\%(\%)   & 20\%(\%)      \\ \hline
			AlphaX~\cite{wang2019alphax}                        & -           & 9.74    & 39.59    & 53.88    & -      & -         \\
			Peephole~\cite{deng2017peephole}                    & 1.64      & -2.35    & 26.49    & 36.52    & 29.42 & 33.31    \\
			RFGIAug(ours)  & \textbf{70.02$\pm$0.14}      & \textbf{73.56$\pm$0.13}    & \textbf{81.86$\pm$0.08}    & \textbf{83.64$\pm$0.02}    & \textbf{84.52$\pm$0.03} & \textbf{85.93$\pm$0.03}   \\ \hline\hline
			E2EPP~\cite{sun2019surrogate}                      & 53.25      & 65.73    & 70.14    & 72.10    & 73.02 & 74.17    \\ 
			E2EPP+GIAug(ours)                 & \textbf{72.10$\pm$0.45}      & \textbf{76.06$\pm$0.23}    & \textbf{77.97$\pm$0.18}    & \textbf{78.72$\pm$0.08}    & \textbf{79.26$\pm$0.34} & \textbf{79.86$\pm$0.09}\\ \hline
			NPNAS~\cite{wen2020neural}        & 64.19& 75.50	& 76.79			& 77.85 &  79.00&  79.43     \\ 
			NPNAS+GIAug(ours)                 & \textbf{74.51$\pm$1.16}      & \textbf{80.41$\pm$2.34}    & \textbf{84.92$\pm$0.12}    & \textbf{85.16$\pm$0.22}    & \textbf{85.26$\pm$0.37} & \textbf{85.32$\pm$0.27}\\ \hline
		\end{tabular}
	\end{center}\label{result_pp_201_Ktau}
\end{table*}

The experiment results measured by Kendall's Tau are shown in Table~\ref{result_pp_101_ktau}, which are composed of three parts. The first is the comparison against AlphaX, Peephole, and ReNAS. The second and the third are about the E2EPP and NPNAS with and without the use of GIAug. Please note that the symbols of ``-'' in the table implies there is no result publicly reported in the corresponding literature. As can be seen from the first part, RFGIAug wins the peer competitors on all the proportions of training data, except for $1\%$. Furthermore, with the use of GIAug, the performance of E2EPP improves on all the different numbers of training samples. The same superiority can also be investigated from NPNAS with the use of GIAug, the performance of NPNAS also enhance in the different settings of training samples.

The experiment results in terms of the $N@K$ metric are shown in Table~\ref{result_pp_101_NK}. Specifically, in each specified number of training samples, the values of $K$ in the metric are specified as $5$ and $10$ as suggested in~\cite{ning2020generic,chen2021not}. As can be seen from Table~\ref{result_pp_101_NK}, RFGIAug outperforms all competitors. Particularly, when using $0.05\%$ of the training data, the value of $N@K$ obtained by RFGIAug accounts for less than $1\%$ obtained by Peephole. The superiority of RFGIAug becomes significant when using $0.1\%$ of the training data, where the values of RFGIAug and Peephole are $14$ and $17,288$, respectively. Furthermore, GIAug can also enhance the performance of E2EPP and NPNAS with the different number of training data under both settings of $K$. 
\begin{table}
	\caption{The comparisons \textbf{(measured by $N@K$)} of the proposed GIAug method on \textbf{NAS-Bench-201} against chosen performance predictors using training data with proportions of $1\%$ and $10\%$ with $K=5$ and $K=10$.}
	\begin{center}
		{
		\begin{tabular}{c|cccc}
			\hline
			\multirow{2}{*}{} & \multicolumn{2}{c|}{1\%} & \multicolumn{2}{c}{10\%} \\ \cline{2-5} 
			& N@5     & \multicolumn{1}{c|}{N@10}    & N@5       & N@10     \\ \hline
			AlphaX~\cite{wang2019alphax}                        & -       & \multicolumn{1}{c|}{-}       & 1538      & 224      \\
			Peephole~\cite{deng2017peephole}                   & 1607     & \multicolumn{1}{c|}{1569}     & 9981       & 529        \\
			RFGIAug(Ours)                       & \textbf{7}      & \multicolumn{1}{c|}{\textbf{4}}      & \textbf{1}         & \textbf{1}         \\ \hline\hline
			E2EPP~\cite{sun2019surrogate}                      & 22     & \multicolumn{1}{c|}{1}     & 1         & 1        \\
			E2EPP+GIAug(Ours)                 & \textbf{1}     & \multicolumn{1}{c|}{\textbf{1}}     & \textbf{1}         & \textbf{1}        \\ 
			\hline\hline
			NPNAS~\cite{wen2020neural}                       &166   & \multicolumn{1}{c|}{166}     & 159      & 3     \\
			NPNAS+GIAug(Ours)                 & \textbf{2}     & \multicolumn{1}{c|}{\textbf{1}}     & \textbf{1}         & \textbf{1} \\  \hline
		\end{tabular}}
	\end{center}\label{result_pp_201_NK}
\end{table}

\subsubsection{\textbf{Comparisons of Predictors on NAS-Bench-201}}\label{sec_comp_pp_201}
Similar to the experiments on NAS-Bench-101, we use the same regression model for implementing the proposed GIAug method and the same peer competitors for the experiments conducted on NAS-Bench-201, except that the ReNAS is not chosen because there are no corresponding results reported in its seminal paper nor source code publicly available. Furthermore, although the proportions of the training data used are different from those specified for NAS-Bench-101, their particular numbers of training data are nearly the same because NAS-Bench-201 has quite fewer training data than NAS-Bench-101 does.

The experiment results measured by Kendall's Tau are shown in Table~\ref{result_pp_201_Ktau}. As can be seen from the second row of this table, RFGIAug still achieves the best performance with substantial improvement on the six different settings regarding the number of training data. This can also be observed from the comparisons against E2EPP and NPNAS. Especially, when E2EPP and NPNAS are implemented with GIAug, their performance is almost kept unchanged when the number of training data is more than $5\%$. This indicates that GIAug is more suitable to the real scenario of utilizing performance predictors, where there are often few labeled architectures. In addition, with much fewer training data, e.g., $0.5\%$, GIAug increases the Kendall's Tau of E2EPP by $35\%$, and NPNAS by $16\%$, which are significant improvements.

The results measured by $N@K$ is shown in Table~\ref{result_pp_201_NK}. As can be seen, RFGIAug wins AlphaX and Peephole in terms of both settings of $k$ upon different numbers of training samples. Specifically, the values of RFGIAug are all one when using $10\%$ training samples, which indicates that RFGIAug has found the architecture with the best performance, and can completely guide the search of the corresponding NAS algorithm. Furthermore, GIAug can also enhance the performance of E2EPP. Specifically, GIAug improves the value of $N@K$ of E2EPP by $21$ places when the proportion is $1\%$ and $K$ equals $5$. A similar superiority can also be observed upon NPNAS. Specifically, when the proportions are $1\%$, the $N@5$ and $N@10$ of NPNAS are improved by $164$ places and $165$ places, respectively. Besides, GIAug improves the $N@5$ and the $N@10$ to $1$ when the proportion is $10\%$.

\subsection{Comparisons against Neural Architecture Search (NAS)}\label{sec4_com_nas}
Although the effectiveness of GIAug has been verified by the comparisons against the state-of-the-art performance predictors, the ultimate purpose of designing GIAug is to improve the performance of NAS algorithms. As a result, the comparisons against state-of-the-art NAS are also conducted in this section. Specifically, the experiments are conducted on two commonly used datasets: CIFAR-10 and ImageNet. The architectures are tested on CIFAR-10 when searching on NAS-Bench-101 and NAS-Bench-201, and ImageNet when searching on DARTS search space. Please note that the experiment results are measured by different combinations of evaluation metrics, for the purpose of keeping consistent with the NAS algorithms compared. Specifically, the test accuracy (in short named test) is used by all comparisons, while for other metrics, we will detail them in the related comparisons. As for GIAug, we still use the RF as the regression model, and then implement a simple NAS based on the standard genetic algorithm to search for the best architecture.

\subsubsection{\textbf{NAS on NAS-Bench-101}}\label{sec_comp_nas_101}
In this experiment, the peer competitors are the corresponding NAS algorithms using the performance predictors of Peephole, E2EPP, ReNAS, and SSANA~\cite{tang2020semi}, and their results used for comparisons are from references~\cite{xu2021renas,tang2020semi}. For the convenience of the discussion, we still use their names to represent the corresponding NAS algorithms for the comparison. The experiments are carried out on $424$ training samples and $1,000$ training samples to train performance predictors, following the conventions in~\cite{xu2021renas,tang2020semi}. In addition, we also employ the ``rank'' as the additional metric, which reflects the true ranking of the searched architecture in the NAS-Bench-101 dataset, considering its popularity in recent works~\cite{xu2021renas,tang2020semi}. Furthermore, we independently perform each NAS algorithm for $20$ runs and report the best results for the comparison. 

\begin{table}
	\caption{Test accuracy rate and ranking of the searched architectures on \textbf{NAS-Bench-101} with $424$ and $1,000$ annotated architectures randomly selected to train predictors.}
	\begin{center}
	{
			\begin{tabular}{c|cc|cc}
				\hline
				\multirow{2}{*}{} & \multicolumn{2}{c|}{424 samples}     & \multicolumn{2}{c}{1,000 samples}       \\ \cline{2-5} 
				& test(\%)        & rank(\%) & test(\%)        & rank(\%) \\ \hline
				Peephole~\cite{deng2017peephole}                & 92.94 & 12.32       & 93.75 & 1.64        \\
				E2EPP~\cite{sun2019surrogate}                   & 93.91 & 1.23        & 93.90 & 0.15        \\
				ReNAS~\cite{xu2021renas}                   & 94.06 & 0.02        & -              & -           \\
				SSANA~\cite{tang2020semi}                   & -              & -           & 94.13 & 0.01        \\ \hline
				RFGIAug                    & \textbf{94.23} & \textbf{0.007}       & \textbf{94.20} & \textbf{0.004}       \\ \hline
		\end{tabular}}
	\end{center}\label{result_nas_101}
\end{table}

The results of the experiment are shown in Table~\ref{result_nas_101}. As can be seen, when using $424$ samples for training the corresponding performance predictors, RFGIAug obtains the highest test accuracy of $94.23\%$, while Peephole, E2EPP, and ReNAS are $92.94\%$, $93.91\%$, and $94.06\%$, respectively. In terms of the metric of ``rank'', RFGIAug achieves the value of $0.007\%$, which is one, three, and four orders of magnitude less than Peephole, E2EPP, and ReNAS, respectively. When the number of training samples increases to $1,000$, RFGIAug is still the champion, having the highest test accuracy and the least value measured by the ``rank'' metric.

\subsubsection{\textbf{NAS on NAS-Bench-201}}\label{sec_comp_nas_201}

Most of the peer competitors in this experiment are the NAS algorithms, which are mainly collected from two different categories. The first is from the seminal paper of NAS-Bench-201 and the second is from the NAS literature. In particular, these algorithms include REA~\cite{real2019regularized} and NPENAS~\cite{wei2020npenas} based on evolutionary computation, REINFORCE~\cite{williams1992simple} and ENAS~\cite{pham2018efficient} based on reinforcement learning, and DARTS-V1~\cite{liu2018darts}, DARTS-V2~\cite{liu2018darts}, GDAS~\cite{dong2019searching}, and SETN~\cite{dong2019one} based on gradient-descent algorithms. Furthermore, we also chose the RSPS~\cite{li2020random} algorithm and the BOHB~\cite{falkner2018bohb} algorithm, which are achieved by random search and hyperparameter optimization, and other three performance predictor-based NAS algorithms (i.e., ReNAS~\cite{xu2021renas}, E2EPP~\cite{sun2019surrogate}, NASBOT~\cite{white2020study}), as the peer competitors.

\begin{table}
	\caption{Validation accuracy and test accuracy of the searched architectures on \textbf{NAS-Bench-201} for \textbf{CIFAR-10}.}
	\begin{center}
		\begin{tabular}{c|c|cc}
			\hline
			    & Val(\%)     & \multicolumn{1}{c|}{Test(\%)}                    & seconds \\ \hline
			RSPS~\cite{li2020random}      & 85.85 & \multicolumn{1}{c|}{91.04}          & 7587.12        \\
			DARTS-V1~\cite{liu2018darts}  & 39.77 & \multicolumn{1}{c|}{54.30}          & 10889.87       \\
			DARTS-V2~\cite{liu2018darts}  & 39.77 & \multicolumn{1}{c|}{54.30}          & 29901.67       \\
			GDAS~\cite{dong2019searching}      & 90.21 & \multicolumn{1}{c|}{93.64}          & 28925.91       \\
			SETN~\cite{dong2019one}      & 87.42 & \multicolumn{1}{c|}{90.82}          & 31009.81       \\
			ENAS~\cite{pham2018efficient}      & 39.77 & \multicolumn{1}{c|}{54.30}          & 13314.51       \\ \hline
			NPENAS~\cite{wei2020npenas}    & 91.19 & \multicolumn{1}{c|}{91.68}          & -              \\
			REA~\cite{real2019regularized}       & 91.50 & \multicolumn{1}{c|}{94.22}          & 0.02           \\
			NASBOT~\cite{white2020study}    & -              & \multicolumn{1}{c|}{93.87}          & -              \\
			REINFORCE~\cite{williams1992simple} & 91.46 & \multicolumn{1}{c|}{94.22}          & 0.12           \\
			BOHB~\cite{falkner2018bohb}      & 91.35 & \multicolumn{1}{c|}{94.13}          & 3.59           \\
			ReNAS~\cite{xu2021renas}     & 91.21 & \multicolumn{1}{c|}{94.24}          & 86.31          \\
			E2EPP~\cite{sun2019surrogate}     & ~\textbf{91.50} & \multicolumn{1}{c|}{94.14}          & 0.06           \\ \hline
			RFGIAug      & 91.43 & \multicolumn{1}{c|}{\textbf{94.25}} & 0.06           \\ \hline
		\end{tabular}
	\end{center}\label{result_nas_201}
\end{table}

In addition to the test accuracy, we also report the validation (denoted as val) accuracy and the query time (measured by seconds) for the comparison, which is still based on the available results publicly made by the corresponding literature. Furthermore, $424$ training samples were randomly selected to train RFGIAug, which is the least number among the performance predictor-based peer competitors, and then the best one is picked up for the comparisons after independently repeating $20$ times.

The experiment results are shown in Table~\ref{result_nas_201}. Specifically, RFGIAug outperforms all 13 peer competitors in terms of test accuracy. In addition, although RFGIAug is slightly inferior to REA, REINFORCE, and E2EPP ($0.07\%$, $0.03\%$, and $0.07\%$, respectively), it wins all the remaining 10 peer competitors with the least improvement of $0.08\%$ (BOHB) and the highest improvement of $59.66\%$ (on both DARTS-V1 and DARTS-V2). Furthermore, RFGIAug consumes significantly less query time compared with the peer competitors collected from the seminal paper of NAS-Bench-201, and the competitive query time compared with the algorithms falling into the second category of the chosen peer competitors.

\subsubsection{\textbf{NAS on DARTS}}\label{sec_comp_nas_darts}
The target dataset used in this group of experiments is ImageNet. The peer competitors chosen are the architecture obtained from NASNet-A~\cite{zoph2018learning}, the architecture searched by AmoebaNet-B~\cite{real2019regularized}, and the architecture reported by DARTS, which are all state of arts. 

\begin{table}[htp]
	\caption{The experiment results on the search space of \textbf{DARTS} for \textbf{ImageNet}.}
	\begin{center}
		\begin{tabular}{c|ccc}
			\hline& GPU days
			& \#Parameter & Test(\%)  \\ \hline
			AmoebaNet-B~\cite{real2019regularized} & 3,150 & 5.3M   & 74.0             \\
			NASNet-A~\cite{zoph2018learning}   & 2,000  & 5.3M   & 74.0             \\
			DARTS~\cite{liu2018darts}     & 4    & 4.7M   & 73.3            \\ \hline
			RFGIAug    & \textbf{1.5} & 4.8M   & 73.4            \\ \hline
		\end{tabular}
	\end{center}\label{result_nas_darts}
\end{table}

Different from the experiments on NAS-Bench-101 and NAS-Bench-201, there are no available labeled architectures sampled from the search space of DARTS, which are used to train GIAug in this experiment. To achieve this, we randomly sampled $100$ architectures from the search space of DARTS, and then trained them one by one to construct an annotated architecture dataset. Specifically, each architecture was trained on CIFAR-10 for $50$ epochs by using the stochastic gradient descent (SGD) algorithm, with an initial learning rate of $0.05$, a momentum of $0.9$, a weight decay of $3.0\times10^{-4}$, and a mini-batch size of $64$. The softmax cross-entropy was used as the loss function. This process consumed $1.5$ GPU days with four NVIDIA 2080TI GPU cards. After the training, these $100$ annotated architectures were used to train GIAug where the RF was still as the regression model. Since DARTS has two different cells, i.e., normal cell and reduction cell, the regression models for these two cells are trained separately. After that, the trained two regression models generated a final model through the model ensemble. With the simply implemented NAS algorithm with GIAug, the best performing cell searched on the CIFAR-10 dataset was then transferred to ImageNet for reporting the experiment results. Please note that this process of transferring is a convention of investigating the performance of NAS algorithms on challenging datasets among the NAS community~\cite{liu2018darts}. Specifically, the architecture was trained on ImageNet for $250$ epochs at an initial learning rate of $0.025$ and a weight decay of $5.0\times 10^{-4}$ with SGD, where the momentum, the loss function, and the mini-batch size were specified as the same as making the annotated architectures above. By following the conventions, the number of parameters regarding the architecture and the GPU day used to obtain the architecture are also used as the metrics in this comparison, in addition to the test accuracy.

The experiment results are reported in Table~\ref{result_nas_darts}. As can be seen from Table~\ref{result_nas_darts}, RFGIAug consumes much fewer GPU days, while still achieving promising test accuracy on ImageNet and also has fewer parameters of the searched architecture. Specifically, RFGIAug uses 1.5 GPU days, while AmoebaNet-B, NASNet-A, and DARTS consume $3,150$ GPU days, $2,000$ GPU days and $4$ GPU days, respectively. In terms of the parameter number, the architecture searched by RFGIAug is slightly larger than that searched by DARTS (4.8M v.s. 4.7M), which are fewer than those searched by NASNet-A and AmoebaNet-B (both are 5.3M). This is in contrast to the test accuracy, where RFGIAug is better than DARTS ($73.4\%$ v.s. $73.3\%$), but inferior to NASNet-A and AmoebaNet-B (both are $74.0\%$). This experiment verifies the effectiveness of the proposed algorithm that primarily aims at improving the efficiency of NAS algorithms without significant performance deterioration based on the non-free-lunch theory. 

\begin{table*}[]
	\caption{Comparisons \textbf{(measured by Kendall's Tau)} on regression models on \textbf{NAS-Bench-101}. 0.1\% proportions architectures of NAS-Bench-101 are used to train the regression models.}
	\begin{center}
		\begin{tabular}{c|cccccccccc}
			\hline
			\multirow{2}{*}{Case} & \multicolumn{9}{c}{Regression Model (with 424 data)}                                                                    \\ \cline{2-10} 
			& SVR    & MLP    & LR     & KNN    & ExtraTree & Bagging & DT     & GBRT   & RF     \\ \hline
			1                     & 0.3152 & 0.0477 & 0.3262 & 0.3353 & 0.3339    & 0.4535  & 0.3297 & 0.4939 & 0.5054 \\
			2                     & 0.3393 & 0.2046 & 0.4916 & 0.4473 & 0.3951    & 0.4904  & 0.3949 & 0.5128 & 0.5151 \\
			3                     & 0.4266 & 0.3719 & 0.3083 & 0.4045 & 0.5046    & 0.6201  & 0.5119 & 0.5687 & 0.6597 \\
			4                     & \textbf{0.4341} & \textbf{0.5108} & \textbf{0.5216} & \textbf{0.5094} & \textbf{0.5195}    & \textbf{0.6397}  & \textbf{0.5245} & \textbf{0.6024} & \textbf{0.6727} \\ \hline
		\end{tabular}
	\end{center}\label{result_abl_101}
\end{table*}

\begin{table*}[]
	\caption{Comparisons \textbf{(measured by Kendall's Tau)} on regression models on \textbf{NAS-Bench-201}. 5\% proportions (781) architectures of training data are used to train the regression models.}
	\begin{center}
		\begin{tabular}{c|ccccccccc}
			\hline
			\multirow{2}{*}{Case} & \multicolumn{9}{c}{Regression Model (with 781 data)}                                                  \\ \cline{2-10} 
			& SVR    & MLP    & LR     & KNN    & ExtraTree & Bagging & DT     & GBRT   & RF     \\ \hline
			1                     & 0.4289 & 0.1717 & 0.4059 & 0.5932 & 0.6760    & 0.7313  & 0.6728 & 0.7200 & 0.7318 \\
			2                     & 0.5576 & 0.3147 & 0.6539 & 0.6550 & 0.6841    & 0.7247  & 0.6793 & \textbf{0.7381} & 0.7334 \\
			3                     & 0.5718 & 0.7889 & 0.4092 & 0.5929 & 0.7121    & 0.7806  & 0.7281 & 0.6832 & 0.7867 \\
			4                     & \textbf{0.6525} & \textbf{0.7890} & \textbf{0.6851} & \textbf{0.7316} & \textbf{0.7681}    & \textbf{0.8067}  & \textbf{0.7643} & 0.7087 & \textbf{0.8189} \\ \hline
		\end{tabular}
	\end{center}\label{result_abl_201}
\end{table*}

\subsection{Ablation Study}\label{exp_ablation_study}
As highlighted in Section~\ref{sec3}, the proposed encoding method and the architecture augmentation method are the core designs of the proposed GIAug method. To demonstrate the effectiveness of both components, the ablation study is conducted in this section. 

To illustrate the capability of the proposed encoding method, the encoding method defined in ReNAS, which has been popularly used by existing algorithms, is chosen as the peer competitor. Specifically, the encoding method of ReNAS uses categorical integers to represent the operation type of vertices. Then, the type vector is broadcasted into the adjacency matrix $M_{A}$ to encode the whole architecture, which can be expressed by Equation~(\ref{eq_renas}):
\begin{equation}\label{eq_renas}
	E(A,m) = M_{A} \times strech(integer(m))\}
\end{equation}
where $A$ denotes the adjacency matrix of the given architecture, $m$ denotes the attribute vector and each operation is an integer in $m$ $integer(\cdot)$, and $strech(\cdot)$ stretches a vector into a square matrix.

To further illustrate the generalizability of the proposed GIAug method, multiple popular regression models are also chosen for the study. These regression models are Decision Tree (DT)~\cite{breiman2017classification}, Linear Regression(LR), Support Vector Regression (SVR)~\cite{chang2011libsvm}, K-Nearest Neighbors (KNN), Random Forest (RF)~\cite{breiman2001random}, Gradient Boosted Regression Tree (GBRT)~\cite{friedman2001greedy}, Bagging~\cite{breiman1996bagging}, and ExtraTree~\cite{geurts2006extremely}. We elaborate the four following cases for the comprehensive ablation study:

$\textbf{case 1 (baseline)}$: Using the encoding method of ReNAS and without the proposed data augmentation;

$\textbf{case 2 (encoding method)}$: Using the proposed encoding method and without the proposed data augmentation;

$\textbf{case 3 (augmentation method)}$: Using the encoding method of ReNAS and the proposed augmentation method;

$\textbf{case 4 (augmentation method + encoding method)}$: Using the proposed encoding method and the proposed augmentation method.

This study is performed on both NAS-Bench-101 and NAS-Bench-201, and Kendall's Tau metric is used for measuring the experiment results. Specifically, the proportions of $0.1\%$ and $5\%$ architectures of NAS-Bench-101 and NAS-Bench-201 are randomly selected as the training data for training the regression models (resulting in $424$ and $781$ architectures, respectively), and the reminding for the testing. For \textbf{case 3} and \textbf{case 4}, where the proposed augmentation method is used, there will be $424\times(7-2)!=50,880 $ and $781\times 6!=562,320 $ architectures augmented on NAS-Bench-101 and NAS-Bench-201, respectively. This is because NAS-Bench-101 is based on ONN, and each architecture has five intermediate vertices (exclusive the input and the output vertice), and NAS-Bench-201 is based on OOE, and each architecture has six intermediate edges.

The experiment results on NAS-Bench-101 are shown in Table~\ref{result_abl_101}. As can be seen, all of the nine different regression models investigated in both cases can benefit from the components of the proposed two designs. Specifically, the Kendall's Tau of \textbf{case 2} outperforms that of \textbf{case 1} and the Kendall's Tau of \textbf{case 4} outperforms that of \textbf{case 3}, which reveals the effectiveness of the proposed encoding method. Besides, \textbf{case 3} wins \textbf{case 1} except for the comparison regarding using RF as the regression model, and \textbf{case 4} wins \textbf{case 2} upon all regression models chosen in this study. This demonstrated that the proposed augmentation method can improve the performance of the performance predictors. Furthermore, the situation with both the encoding method and the augmentation method gain the best value of Kendall's Tau, as highlighted in the last row of Table~\ref{result_abl_101}.

The results of the experiments measured by Kendall's Tau on NAS-Bench-201 are shown in Table~\ref{result_abl_201}, which shows similar results to those obtained on NAS-Bench-101. In particular, the encoding method is proved to be effective as can be seen in the comparison between \textbf{case 1} and \textbf{case 2} and the comparison between \textbf{case 3} and \textbf{case 4}. Only in the comparison between \textbf{case 1} and \textbf{case 2} regarding using Bagging as the regression model, the Kendall's Tau of \textbf{case 2} is slightly lower than that of \textbf{case 1}. In addition, \textbf{case 3} wins \textbf{case 1} except for the comparison in terms of KNN and GBRT, and \textbf{case 4} won \textbf{case 2} except for the comparison of GBRT. Furthermore, \textbf{case 4} with the whole GIAug (i.e., both encoding method and the augmentation method) wins other cases except for the comparison in terms of using GBRT as the regression model.

In summary, these $72$ experiment results of the ablation study, conducted on two different datasets with nine different regression models, demonstrate the respective superiority of both designs. In addition, these results also demonstrate the effectiveness of collectively using both designs. Please note, the best results on both datasets are all obtained in terms of using the RF as the regression model. This is also the reason for selecting RF in other experiments.

\section{Conclusions}\label{sec_5}
The goal of this paper was to develop a method to solve the shortage of training data faced by performance predictors. This goal has been achieved by proposing an effective architecture augmentation method (named GIAug). GIAug mainly has three advantages: computational cheapness, applicable generality and superior effectiveness. Specifically, the proposed augmentation method is based on the isomorphism mechanism, which can generate many high-quality DNN architectures labeled yet with negligible computation resources consumed. In addition, a generic encoding method has also been utilized. This made the proposed GIAug method can be embedded in various performance predictors. Furthermore, GIAug was first examined against state-of-the-art performance predictors with various proportions of data from the benchmarks. The results demonstrated the effectiveness of GIAug among comparisons. Finally, GIAug was also investigated against state-of-the-art NAS algorithms on three popular search spaces, and the performance gained by GIAug was also very competitive on different scales of image classification datasets.
With the development of NAS, more and more related work is being proposed for other types of DNNs, such as the recurrent neural network and graph neural network. In the future, we will extend our work for these types of architectures.

\ifCLASSOPTIONcaptionsoff
  \newpage
\fi

\bibliographystyle{IEEEtran}
\bibliography{reference}

\end{document}